\documentclass[conference]{IEEEtran}
\IEEEoverridecommandlockouts
\usepackage{bm}
\usepackage{natbib}
\usepackage{multirow}
\usepackage{amsmath,amssymb,amsfonts}
\usepackage{algorithmic}
\usepackage{graphicx}
\usepackage{textcomp}
\usepackage{xcolor}
\usepackage{indentfirst}
\usepackage{float}
\usepackage{amsthm}
\theoremstyle{definition}
\setcitestyle{square,comma,numbers}
\newtheorem{defn}{Definition}
\def\BibTeX{{\rm B\kern-.05em{\sc i\kern-.025em b}\kern-.08em
    T\kern-.1667em\lower.7ex\hbox{E}\kern-.125emX}}
\begin{document}

\title{Inference for Network Structure and Dynamics from Time Series Data via Graph Neural Network}

\author{\IEEEauthorblockN{Mengyuan Chen}
\IEEEauthorblockA{\textit{School of Systems Science} \\
\textit{Beijing Normal University}\\
BeiJing, China \\
chenmy@mail.bnu.edu.cn}
\and
\IEEEauthorblockN{Jiang Zhang*}
\IEEEauthorblockA{\textit{School of Systems Science} \\
\textit{Beijing Normal University}\\
BeiJing, China \\
zhangjiang@bnu.edu.cn}
\and
\IEEEauthorblockN{Zhang Zhang}
\IEEEauthorblockA{\textit{School of Systems Science} \\
\textit{Beijing Normal University}\\
BeiJing, China \\
3riccczz@gmail.com}
\and
\IEEEauthorblockN{Lun Du}
\IEEEauthorblockA{\textit{Data Knowledge Intelligence Group}\\
\textit{Microsoft Research}\\
BeiJing, China \\
lun.du@microsoft.com}
\and
\IEEEauthorblockN{Qiao Hu}
\IEEEauthorblockA{\textit{Technology Group} \\
\textit{Swarma Campus (Beijing) Technology}\\
BeiJing, China\\
huqiao@swarma.org}
\and
\IEEEauthorblockN{Shuo Wang}
\IEEEauthorblockA{\textit{School of Systems Scienc} \\
\textit{Beijing Normal University}\\
BeiJing, China \\
shawnwang.tech@gmail.com}
\and
\IEEEauthorblockN{Jiaqi Zhu}
\IEEEauthorblockA{\textit{School of Systems Scienc} \\
\textit{Beijing Normal University}\\
BeiJing, China \\
zhujiaqi@mail.bnu.edu.cn}
}

\maketitle

\begin{abstract}
Network structures in various backgrounds play important roles in social, technological, and biological systems. However, the observable network structures in real cases are often incomplete or unavailable due to the measurement errors or private protection issues. Therefore, inferring the complete network structure is useful for understanding complex systems. The existing studies have not fully solved the problem of inferring network structure with partial or no information about connections or nodes. In this paper, we tackle the problem by utilizing time series data generated by network dynamics. We regard the network inference problem based on dynamical time series data as a problem of minimizing errors for predicting future states and proposed a novel data-driven deep learning model called Gumbel Graph Network (GGN) to solve the two kinds of network inference problems: Network Reconstruction and Network Completion. For the network reconstruction problem, the GGN framework includes two modules: the dynamics learner and the network generator. For the network completion problem, GGN adds a new module called the States Learner to infer missing parts of the network. We carried out experiments on discrete and continuous time series data.  The experiments show that our method can reconstruct up to 100\% network structure on the network reconstruction task. While the model can also infer the unknown parts of the structure with up to 90\% accuracy when some nodes are missing. And the accuracy decays with the increase of the fractions of missing nodes. Our framework may have wide application areas where the network structure is hard to obtained and the time series data is rich.
\end{abstract}

\begin{IEEEkeywords}
Network Inference, Network Reconstruction, Network Completion, Graph Network, Time Series
\end{IEEEkeywords}

\section{Introduction}
A complex system is an integrated system with many parts, and the emergent behaviors of the entire system are determined by how these parts interact \cite{i5}, i.e., the network structure. For example, the topology of a social network determines how fast the opinions or ideas can spread in a social media \cite{i1,i2}; the structure of a supply chain network between companies influences the safety of the whole market because risk may propagate along the links \cite{klibi2012scenario,cimini2015systemic}; the topology of the cooperation network plays critical role for scientific innovation and individual development for young scientists \cite{newman2001scientific,i3}. However, the data of network structure is always incomplete or even unavailable either because measuring binary links is costly or the data of weak ties is missing \cite{Kossinet2006effects,anand2018missing,cimini2015systemic}. Therefore, it is urgent to find a way to infer the complete network structure according to non-structural information \cite{squartini2018reconstruction,guimera2009missing}.

Link prediction, as the traditional task in network science, tries to infer the lost links in network structure according to the linking patterns of existing connections \cite{kunegis2009learning,lu2009similarity}. Although numerous algorithms have been developed to complete the missing links of a large network with high accuracy \cite{wang2011human,zhang2018link}, all of these approaches require the complete nodes information but it is always unavailable in practice \cite{Kossinet2006effects,tran2019deepnc}. link prediction cannot solve the inference problem under the condition that the network contains unobservable nodes. In real cases, we can either obtain nodes information of the only partial network or without any information about links \cite{guthke2004dynamic,tran2017community}, as a result, conventional link prediction algorithms cannot work.

Network completion methods have been developed in recent years trying to tackle the first problem we will discuss in the paper, that is, to infer the missing connections on unobserved nodes according to the linking patterns between observable nodes. The methods can be categorized into traditional statistical-based methods and graph neural network liked methods. Give examples of traditional methods represented by expectation maximum(EM) algorithm, Myunghwan Kim and Jure Leskovec applied Kronecker graph model and EM to complete the network according to the observed linking patterns \cite{kim2011network}. Although their algorithms obtain a relatively high accuracy of recovering missing links, there is an implicit requirement, the underlying network structure is required to follow the self-similar property as possible, which is violated by some network \cite{leskovec2010kronecker}. Following the same expectation maximum(EM) algorithm, in a recent NC work, Xue, Yuankun and Bogdan, Paul developed a causal inference method to recover the complete network structure from the adversarial interventions \cite{xue2019reconstructing}. 
On the other hand, as the booming development of deep learning on graphs \cite{wu2019comprehensive,du2018dynamic,du2018galaxy}, researchers applied graph convolution network (GCN) liked models on network completion problems. Da Xu et al regard the complete network as the growth of the partial network 
\cite{xu2019generative}, then they train the GCN to learn the growing process with the partially observed network and generalized to the complete unknown network structure. Cong Tran, Won-Yong Shin, and Andreas Spitz et al solve the problem by training a graph generating model to learn the connection patterns among a large set of similar graphs for training and applied the trained generating model to complete the missing information \cite{tran2019deepnc}. All of these network completion methods depend on a partially observed network structure because they try to discover the latent patterns of the observed connections and to infer the unknown structures. Nevertheless, in some cases the network structures are totally implicit and only some signals of partial nodes can be observed, such as biological network \cite{geier2007reconstructing} and social network \cite{Kossinet2006effects}. How can we infer the whole network structure without any information of connection patterns?

In fact, time series data on nodes as another important information source \cite{krajec2016highlighting,lin2018variational}, which is always available in practice, is ignored by mentioned previous works. For example, in an online social network, we can only observe the discrete retweet events between a large set of users, neither their features like sex, education, etc. and their connection information is unavailable; in a stock market, all the information that we can obtain is the prices of different stocks, the connections between the stocks are unknown. Thus, can we develop a method to infer the network structure according to the time series data representing the nodes'states? A large number of methods have been proposed for reconstructing network from time series data. One class of them is based on the method of statistical inference such as Granger causality \cite{brovelli2004beta,quinn2011estimating}, and correlation measurements \cite{stuart2003gene,eguiluz2005scale,barzel2013network}. Even though, these methods may fail to reveal the structural connection. Another class of methods was developed for reconstructing structural connections directly under certain assumptions. For example, methods such as driving response \cite{timme2007revealing} or compressed sensing \cite{wang2011human,wang2011predicting,wang2011network,shen2014reconstructing} either require the functional form of the differential equations, or the target-specific dynamics, or the sparsity of time series data. However, getting this information is very difficult. Thus, a general framework for reconstructing network topology, completing missing structures, and learning dynamics from the time series data of various types of dynamics, including discrete and binary ones, is necessary.

In this paper, we develop a framework for network inference from the time series data. We discuss two kinds of network inference problems. The network reconstruction problem is defined as the whole network structure reconstruction based on observed nodes' states time series data. In this problem, all nodes are observable. The second problem is network completion, in which, only partial nodes' states time series data is available, and we will infer the complete network structure according to this piece of information under the conditions either the connections between observable nodes are known or unknown. Both problems are formulated as the same kind of optimization problem, which is to find an optimized network structure and the approximator of the network dynamics such that the errors between the observed time series and the generated time series according to the candidate network structure and dynamical rules is minimized. Both problems are solved in the same framework called Gumbel-Graph-Network (GGN) \cite{zhang2019general}, which is a combination of the network generator based on Gumbel softmax sampling and the dynamics learner based on Graph Network. Gumbel softmax sampling is a technique to simulate the sampling process with a differentiable computation process. Equipped with this technique, we can train a network generator with a gradient descent method. Graph Neural Network (GNN) is a new deep learning architecture on graph. By learning a bunch of functions defined on nodes and links representing propagation and aggregation, GNN can recover the complex dynamical process defined on a graph like rigid body movement \cite{kipf2018neural}, coupled oscillators \cite{zhang2019general}, traffic flows \cite{yu2017spatio}, pollution spreading \cite{qi2019hybrid}, etc. By deploying the powerful capability of graph neural network in learning, we can simulate the underlying dynamics based on the real network. 
\section{Network Inference Methodologies}
In this paper, we focus on the network inference problems of the structure and dynamic based on states evolution time series of all or partial nodes. According to the two different application scenarios, we divide the problem into two sub-problems: (1) Network reconstruction problem: inferring interconnected structure and dynamic of network evolution with all nodes observable; (2) Network completion problem: recovering the structure and nodes' states of the entire network based on partial network structure and time series data of the observable nodes. In this section, we first introduce the formal definitions of the two sub-problems and then describe the specific models to solve the problems separately.

Suppose our studied system has an interaction structure described by a binary graph $G = (V, E)$ with an adjacency matrix $A$, where $V = \{v_1, ..., v_N\}$ is the set of nodes, or interchangeably referred to as vertices, and $N$ is the total number of nodes, $E = \{e_{ij}\}$ is the set of edges between the nodes, and $A$ is a binary matrix of which each entry equals $0$ or $1$.

The network dynamic $\mathcal{S}(\psi,A)$ is defined on the graph $G$, where $\psi$ is the dynamical rule which mapping the states of nodes $X(t)=(X_1(t),X_2(t),\cdot\cdot\cdot,X_N(t))\in \mathcal{R}^{n\times d}$ in the system at time $t$ to the states $X(t+1)$ at time $t+1$, where $X(t+1)=\psi(X(t))$, and $d$ is the dimension of the states.
Next, we propose the definition of the network reconstruction problem:

\subsection{Problem Definition}
\begin{figure*}[ht!]
\centering
\includegraphics[scale=0.4]{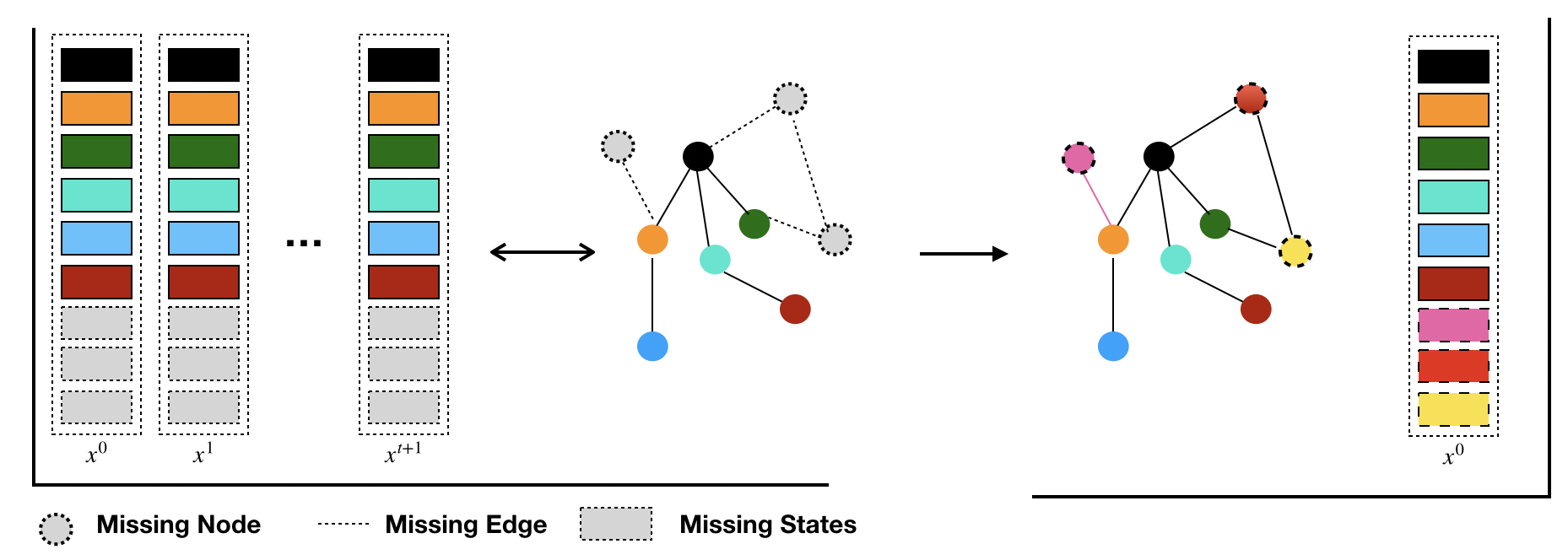}
\caption{The network completion problem on time series : 
the states of some nodes are missing. we only observe the network structure of the remaining nodes (with bold circles), and the aim is to infer the missing information (the dashed circles).}
\label{fig:ncdefinition}
\end{figure*}
\begin{defn}[\textbf{Network Reconstruction}]
The so-called network reconstruction problem refers to deduce the unknown network structure $G$ and dynamic rule $\psi$ from the known nodes' states time series when they can be observed. We can generate a time series with length ${T+1}$ from the system dynamic $\mathcal{S}$, denoted by $\bm{x^{0:T}}=(\bm{x^0},\bm{x^1},\cdot\cdot\cdot,\bm{x^T})$, where $\bm{x^t}=\psi^{t+1}(\bm{x^0)}$, and $\bm{x^0}\in \mathcal{R}^{{\rm S}\times n\times d}$ is the initial states, where $\rm{S}$ is the total number of time series (the number of samples from different initial conditions). The network reconstruction problem is defined as an optimization problem that finds a set of optimal parameters $\alpha,\beta$, to minimize the error value between the state estimation value and ground-truth, which is the objective function formula \ref{loss}.
\begin{equation}
    minimize_{\alpha,\beta}\ L = \sum_{t=1}^{T}||\bm{x^t} - \bm{\hat{x}^t}(\alpha,\beta)||\label{loss}
\end{equation}
such that,
\begin{equation}
    \bm{\hat{x}^t}(\alpha,\beta) = \hat{\psi_{\alpha}}(\bm{x^{t-1}},\hat{A}(\beta)) \label{nr definition}
\end{equation}
for $t=1,2,\cdot\cdot\cdot,T$.

Here, $\hat{\psi_{\alpha}}(\cdot)$ is a dynamical rule parameterized by $\alpha$ to estimate $\psi$. Starting from state \bm{$x^0$}, we iteratively apply it to the states \bm{$x^t$} in the previous steps, and then obtain the estimated evolutionary trajectory $(\bm{\hat{x}^1},\bm{\hat{x}^2},\cdot\cdot\cdot,\bm{\hat{x}^T})$ similar to $(\bm{x^1},\bm{x^2},\cdot\cdot\cdot,\bm{x^T})$; while $\hat{A}(\beta)$ is the estimate of the Adjacency Matrix $A$ with the parameter $\beta$. After minimizing the objective function, from the parameter $\alpha$ we can calculate the optimized dynamical rule and also we can get an estimate of the network structure by sampling from the parameter $\beta$. We hope that $\hat{\mathcal{S}}=(\hat{\psi}(\alpha),\hat{A}(\beta))$ will be close enough to the ground truth $\mathcal{S}$.  
\end{defn}
\begin{defn}[\textbf{Network completion}]
Different from the task of reconstruction, the network completion task is defined in such a case: when part of individuals in the network are unobservable, only the generation time series and network structure of the observable individuals can be observed. As Figure \ref{fig:ncdefinition} described, under these conditions, we still need to infer all the unknown information from the observable individuals, including the dynamical rule, the states information of the unknown nodes, and the structure information between the unknown nodes and the observed nodes.

Let's formulate the problem as follows: we assume that the graph $G = (V,E)$ can be divided into two parts: observed structure $V_o,E_o$ and unobservable structure $V_m,E_m$, accordingly, the adjacency matrix, the states of the nodes, are also divided into two parts $A = A_o\bigoplus A_m$ and $\bm{x} = \bm{x}_o \bigoplus \bm{x}_m$, respectively. Where, $\bigoplus$ indicates the concatenation of tensors with appropriate ways. Network completion problem is then to find a set of $\{\alpha,\beta,\gamma\}$ optimal parameter combinations such that the estimated and the real values of the observed partial time series are as consistent as possible, that is Equation \ref{loss nc}
\begin{equation}
 minimize_{\alpha,\beta,\gamma}\ L = \sum_{t=1}^{T}||\bm{x}_{o}^t- \hat{\bm{x}_{o}}^t(\alpha,\beta,\gamma) ||  \label{loss nc},
\end{equation}
such that:
\begin{equation}
    \hat{\bm{x}^{t}} =  (\hat{\bm{x}_o^t} \bigoplus \hat{\bm{x}_m^t})
\end{equation}
\begin{equation}
    \hat{\bm{x}_o^t} \bigoplus \hat{\bm{x}_m^t} = \hat{\psi_{\alpha}}(\bm{x}_o^{t-1}\bigoplus \hat{\bm{x}_m^{t-1}},A_o\bigoplus\hat{A_m}(\beta)). \label{evolve nc}
\end{equation}
for $t=1,2,\cdot\cdot\cdot,T$. Where $\hat{\psi_{\alpha}}(\cdot)$, $\hat{A_m}(\beta)$ are the estimates of $\psi$ and $A_m$ obtained from the parameters $\alpha,\beta$ respectively. And
\begin{equation}
    \hat{\bm{x}_m}^{t=0} = \rho(\gamma),
\end{equation}
where $\rho(\gamma)\in \mathcal{R}^{S\times M\times d}$ is an estimate of the unknown nodes' initial state parameterized by $\gamma$, where $M$ is the number of unobserved nodes. 
\end{defn}
\subsection{Network reconstruction with Gumbel Graph Network framework}
\begin{figure*}[ht!]
\centering
\includegraphics[scale=0.35]{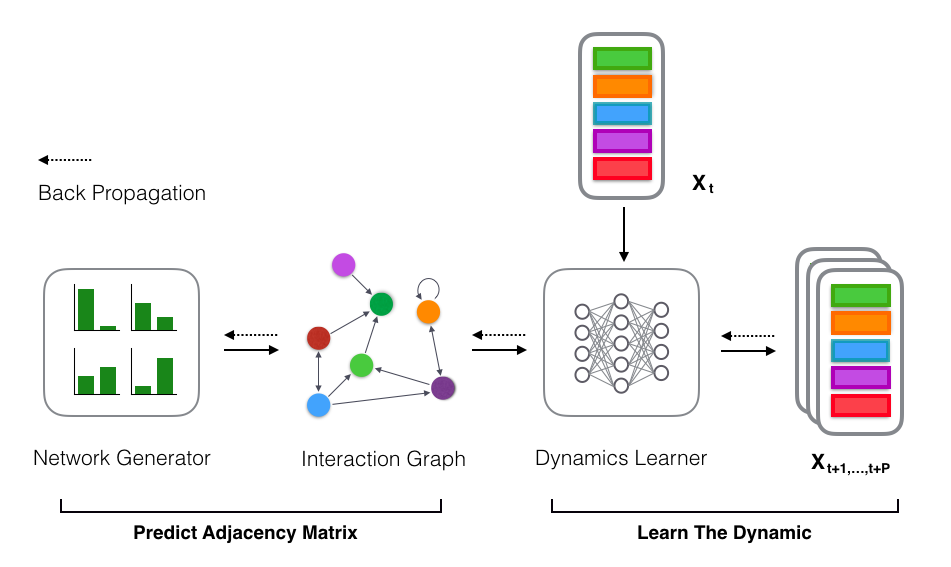}\label{reconstruction framework}

\caption{GGN Modules: The Adjacency Matrix is generated by the Network Generator via Gumbel softmax sampling; then the adjacency matrix and $\bm{x^t}$
(node state at time t) are fed to Dynamic Learner to predict the nodes' states in future P time steps}
\label{fig:reconstruction}
\end{figure*}
To solve both the problems formulated in the previous sections, we extended our previous work, a general deep learning framework called Gumbel Graph Network (GGN) \cite{zhang2019general}. As shown in Figure 2, our general idea is to use a graph network, which is called as a dynamic learner, on the generated candidate network to realize the dynamical rule $\hat{\psi_{\alpha}(\cdot)}$, where, $\alpha$ are learnable weights of the graph network. And the candidate network is generated by a series of gumbel softmax sampling processes parameterized by a matrix $\beta_{N\times N}$, that is,

\begin{equation}
A_{ij} = \frac{\exp((\log(\beta_{ij})+\xi_{ij})/\tau)}{\exp(\log(\beta_{ij})+\xi_{ij})/\tau))+\exp(\log(\beta_{ij})+\xi_{ij}')/\tau))},
\label{eq:gumbel}
\end{equation}
where $\beta_{ij}$ is the probability of connection between node $i$ and node $j$, and $\xi_{ij}$ are i.i.d random variables of the standard Gumbel distribution, and $\tau$ is the temperature parameter. When $\tau$ goes to infinity, $A_{ij}$ will converge to $0$ or $1$. Equation \ref{eq:gumbel} simulates the sampling process of generating $A_{ij}$ with the probability $\xi_{ij}$, however, it is derivable such that it can be adjusted by the gradient descent method.

In the dynamic learner module, we use multiple layered perceptrons (MLP) to simulate the one step of the complex non-linear process in real dynamic $\mathcal{S}$, that is, to complete the mapping from $\bm{x}^{t-1}$ to $\hat{\bm{x}^{t}}$. The module can be replaced by CNN or RNN module.

\begin{equation}
    \hat{\bm{x}^t} =\hat{\psi_{\alpha}}(\hat{A}, \bm{x}^{t-1})
\end{equation}
where, $\alpha$ are the weights of MLPs. The readers can be referred to the paper \cite{zhang2019general} for the details. By contrasting with the ground truth of states, we calculate the gradients of all parameters and update the adjacency matrix parameters $\beta_{ij}$ and $\hat{\psi_{\alpha}}$ module accordingly. 

We alternatively train the network generator and dynamic learner for obtaining the best parameters $\alpha,\beta$.
\begin{table*}[h!]
\centering 
\begin{tabular}{l} 
 \hline
\textbf{Algorithm 1} : NC-GGN algorithm  \\ [0.5ex] 
 \hline
 1 Input: observed adjacency $A_o(V_o,E_o)$; observed states $x_o^0$; \\
Length of Prediction Steps P; Length of Dynamic Learner, Initial States, Gumbel Generator Train Steps D,I,K;\\
 2 Output: predict adjacency $\hat{A}(V_o,E_o,V_m,E_m)$;$\bm{\hat{x}} = \{{\bm{\hat{x}_m^0},\bm{\hat{x}_m^1},\cdot\cdot\cdot,\bm{\hat{x}_m^p},\bm{\hat{x}_o^1},\cdot\cdot\cdot,\bm{\hat{x}_o^p}}\}$  \\
   \# initialization \\
 4 Initialize Dynamics Learner parameters $\alpha$ \\
 5 Initialize Initial States Learner parameters $\gamma$\\ 
 6 Initialize Gumbel Generator parameters $\beta$ \\
   \\
  \# Training \\ 
 7 Initial states of missing nodes $\bm{\hat{x}_m^0} = \rho(\gamma)$ \\
 8 \textbf{for} each epoch \textbf{do} \\
 \hspace*{1.0cm}\# Training Dynamics Learner \\
 9 \hspace*{0.4cm} \textbf{for} \underline{d=1,$\cdot\cdot\cdot$,D} \textbf{do}  \\
 10 \hspace*{0.9cm} $\bm{\hat{x_o}^0} \leftarrow \bm{x_o^0}$ \\
 11 \hspace*{0.9cm} \textbf{for} t=1,$\cdot\cdot\cdot$,P do  \\
 12 \hspace*{1.3cm} $\bm{\hat{{x_o}^t}} \leftarrow$ Dynamics\hspace*{0.1cm} Learner$(A_o,\bm{\hat{x_o}^{t-1}},\alpha)$ \\
 13 \hspace*{0.9cm} \textbf{end} \\
 14 \hspace*{0.9cm} loss $\leftarrow$ Compute\hspace*{0.1cm} Loss$(\{\bm{{x_o}^1},\cdot\cdot\cdot,\bm{{x_o}^p}\},\{\bm{\hat{x_o}^1},\cdot\cdot\cdot,\bm{\hat{x_o}^P}\})$ \\
 15 \hspace*{0.9cm} update $\alpha$ with the gradient of loss \\
 16 \hspace*{0.4cm} \textbf{end} \\
 \\
 \hspace*{1.0cm} \# Training States Learner \\
 17 \hspace*{0.4cm} Missing Edge info:$\hat{A}_m \leftarrow$ Gumbel\hspace*{0.1cm} Generator$(\beta)$ \\
 18 \hspace*{0.4cm} $\hat{A} \leftarrow  (A_o\bigoplus \hat{A}_m)$ \\
 19 \hspace*{0.4cm} \textbf{for} \underline{i=1,$\cdot\cdot\cdot$,I} do \\
 20 \hspace*{0.9cm} $\bm{\hat{x}^0} \leftarrow (\bm{x_o^0} \bigoplus \bm{\hat{x}_m^0})$ \\
 21 \hspace*{0.9cm} \textbf{for} \underline{t=1,$\cdot\cdot\cdot$,P} do  \\
 22 \hspace*{1.5cm} $\bm{\hat{{x}^t}} \leftarrow$ Dynamics\hspace*{0.1cm} Learner$(\hat{A},\bm{\hat{x}^{t-1}},\alpha)$ \\
 23 \hspace*{0.9cm} \textbf{end} \\
 24 \hspace*{0.9cm} loss $\leftarrow$ Compute\hspace*{0.1cm} Loss$(\{\bm{{x_o}^1},\cdot\cdot\cdot,\bm{{x_o}^p}\},\{\bm{\hat{x_o}^1},\cdot\cdot\cdot,\bm{\hat{x_o}^P}\})$ \\
 25 \hspace*{0.9cm} update $\bm{\hat{x}_m^0}$ with the gradient of loss \\
 26 \hspace*{0.4cm} \textbf{end} \\
 \\
 \hspace*{1.0cm} \# Training Network Generator \\
 28 \hspace*{0.4cm} \textbf{for} \underline{k=1,$\cdot\cdot\cdot$,K} do \\
 29 \hspace*{0.9cm} $\bm{\hat{x}^0} \leftarrow (\bm{x_o^0} \bigoplus \bm{{x_m}^0)}$ \\
 30 \hspace*{0.9cm} Missing Edge info:$\bm{\hat{A}_m} \leftarrow$ Gumbel\hspace*{0.1cm} Generator$(\beta)$ \\
31 \hspace*{0.9cm} $\hat{A} \leftarrow$  onehot$(A_o \bigoplus \hat{A}_m)$ \\
32 \hspace*{0.9cm} \textbf{for} \underline{t=1,,P} do  \\
33 \hspace*{1.5cm} $\bm{\hat{{x}^t}} \leftarrow$ Dynamics\hspace*{0.1cm} Learner$(\hat{A},\bm{\hat{x}^{t-1}},\theta)$ \\
34 \hspace*{0.9cm} \textbf{end} \\
35 \hspace*{0.9cm} loss $\leftarrow$ Compute\hspace*{0.1cm} Loss$(\{\bm{{x_o}^1},\cdot\cdot\cdot,\bm{{x_o}^p}\},\{\bm{\hat{x_o}^1},\cdot\cdot\cdot,\bm{\hat{x_o}^P}\})$ \\
36 \hspace*{0.9cm} update $\beta$ with the gradient of loss \\
37 \hspace*{0.4cm} \textbf{end} \\
38 \textbf{end} \\
\hline
\end{tabular}
\label{table:algorithm 1}
\end{table*}
\subsection{Network Completion Gumbel Graph Network}
Similar to the previous subsection, we propose the Network Completion Gumbel Graph Network (NC-GGN) to work out the network completion problem. The inputs of our model are the states \bm{$x_o^t$} of observed nodes and the observed adjacency matrix $A_o$. Correspondingly, the outputs are the complete network structure and the future states of all nodes.

We also use the GGN framework to learn the dynamic and structure. However, the difference is we must additionally learn the state variables of unobserved nodes because all of their states are missing. This will make the problem harder than the network reconstruction problem.

Thus, we designed three modules in our model: (1) The dynamic learner, to predict the states of the node at time $t$ by using the states information at time $t-1$ and the structure information of the observable node $A_o$; (2) The initial state learner, to randomly generate the initial states of the missing nodes $V_m$ with a set of learnable parameters $\gamma$; (3) The network generator, to generate the candidate connections between missing nodes and observable nodes.

In the dynamic learner and network generator modules, we exploited the same techniques like graph network and Gumbel softmax sampling. However, in this way, we can only observe the evolutionary information of known nodes, so we only use the observed node states as supervised information. As to the initial state learning module, the main goal is to find the optimal initial states of the missing nodes by using the same objective function with the optimization method: gradient descent. The generation process of the initial states of unobservable nodes can be formulated in the Equation \ref{miss initial state}:
\begin{equation}
    \hat{\bm{x}_m^0} = \rho(\gamma), \label{miss initial state}
\end{equation}
where $\rho$ is the generation function of initial states parameterized by $\gamma$, and the form of the function depends on the problems. For the simplest case, $\rho$ is just the identical mapping, which means learnable parameters $\gamma$s are the initial states. We can use a stochastic gradient descent algorithm to optimize all the modules with the automatic differential techniques.

To detail the whole process of network completion, we will layout the pseudo-codes of NC-GGN in Algorithm 1.

We separately train the three modules mentioned previously in one epoch, and each module for
multiple rounds. Modules update the parameters within themselves only. That means, for example, only $\alpha$s are updated in dynamic learning modules and keep $\beta$s and $\gamma$s unchanged. 

We start training the dynamic learner for $D$ rounds and the initial state for $I$ rounds in an epoch, and the network generator module for $K$ rounds. We repeat the training process until the model converges or the loss function does not drop.

\section{Experimental Results}
Our framework and algorithms can work on state time series of any format, such as continuous real vectors, discrete tensors, or binary strings. To apply the framework in the scenarios of social science, we construct two examples. 

The first case is the spreading process of opinions. Suppose we can only observe the retweet events from user A to user B with time stamps $t$, and try to reconstruct the social network structure behind the users. The observed retweet events can be converted binary states time series data. Suppose there are $N$ users, and all the users on the heads of the propagating event chains are infected by the opinion. We set the initial states of these sources as $1$ and all other users as $0$. If the propagating event took place at time $t$, from user A to B. Then the state of user B will be converted from $0$ to $1$. And all the users will keep their states unchanged in other cases. In this way, we can convert propagating events into binary states time series. Hence, our algorithms can be applied to reconstruct or complete the network and dynamics.The second example is to predict the prices or volumes of a bunch of stocks. And by using the network reconstruction or completion algorithms, we may also reconstruct the connections between these stocks. The connections may reveal latent information such as joint ownership or economic connections, and facilitate our understanding of the market. Suppose the price or the volume can be described by real values at each time step, then we can obtain the real-valued vectors as the time series. Our algorithms can be applied to this example. However, due to the limitation of data availability and computing resource, we generate data from artificial simulations to test how our algorithms work.

\subsection{Datasets}
To test our framework, we create two data sets by simulations. One data set contains binary states generated by the Voter model, which can simulate the information spreading process on social networks, and the other set contains continuous time series data generated by the Coupled Map Lattice model, which can emulate the fluctuation of the stock price. All simulations are implemented on the small-world networks generated by Watts-Strogatz (WS) model. We will introduce the two simulation models in detail.\\
\textbf{Voter model}. The voter model introduced by Richard A. Holley and Thomas M. Liggett in 1975\cite{holley1975ergodic} can simulate the spreading dynamics of opinions, ideas, information on a network. Suppose there are $N$ interacting agents connected to form a network. Initially, each agent has a distinct ``opinion'' represented by $0$ or $1$. At each time $t$, any agent $i$ will have a chance to change his ''opinion'', and the probability to adopt opinion is determined by the relative fraction of $k$ in all of $i$'s neighbors. 

We generated simulated data on Watts–Strogatz Network with a reconnection probability of $0.2$. We simulate on networks with size 10, 20 or 30 for 200 times (samples) with different initial states, and each simulation runs for 50 steps. Each step in one experiment is a sample of data, so there are 10000 data samples in total. All the 10000 data samples are separated into training, validation and testing data sets with 70\%, 15\%, and 15\% for both network reconstruction task and network completion task.  we randomly removed $M$ nodes and their edges from the WS network with size 10, 20 as observed incomplete graph $G_o$ on the network completion task. 

\textbf{Coupled map lattices}. A Coupled map lattices (CML) model is a dynamical system with discrete time, discrete space, and continuous state variables proposed by Kaneko in 1992 \cite{kaneko1992overview}. We suppose the CML model can generate chaotic time series which can emulate stock price fluctuations. Each element on a lattice consists of a logistic map coupled to their neighbors, this can be written as 
\begin{equation}
    x_{t+1}(i) = (1-\epsilon)f(x_{t}(i))+\frac{\epsilon}{|N_i|}\sum_{j\in {N_i}}{x_{t}(i)}f(x_{t}(j))
\end{equation}
where $x_{t}(i)$ is treated as the state of the element of node $i$ at time $t$, $N_i$ represents node $i$'s neighbors, $\epsilon$ is the coupling constant which can tune the system behavior to the chaos. And as the local map $f(x)$, it usually takes the logistic map:
\begin{equation}
    f(x) = x(1-x)
\end{equation}
For network reconstruct task, we generated simulated data on Watts–Strogatz Network with a reconnection probability of $0.2$.
We run the simulation for 5000 times (samples) with different initial states on networks with different sizes 10, 20 or 30, respectively. For network completion task, we generated simulated data on the same network structure as reconstruct task. We run the simulation for 2000 and 6000 times (samples) with different initial states on networks with different sizes 10 or 20, respectively. For both tasks, each simulation runs for 100 steps. We group every 10 steps as one data record. So there are 50000 data records in total at each size on the network reconstruct task, and 20,000 and 60,000 data records for 10 or 20 sized networks respectively on the network completion task. All the data samples are separated into training, validation and testing data sets with 70\%, 15\% and 15\%.  we randomly removed $M$ nodes and their edges from the WS network as observed incomplete graph $G_o$ on the network completion task. 

\subsection{Performance Metrics}
On the network inference task, we evaluate the performance of our model mainly by the accuracy of nodes' states prediction and the accuracy of network structure prediction. The accuracy of nodes' states is calculated by mean absolute error (MAE). The accuracy of network structure can be done by comparing the adjacency matrices between the prediction value and the ground truth, which can be regarded as a binary classification problem. Therefore, the evaluation index of the binary classification problem like AUC, ACC, TPR and so forth are used to measure the performance of network inference. The measures we used are listed in the following items:
\begin{itemize}
\item[-] \textbf{MAE}(mean absolute error): MAE is a measure of the difference between two vectors, it is the total absolute value of the differences.

\item[-] \textbf{AUC}(area under the roc curve): AUC evaluates how similar the probabilities of network edges from the network generator are to the real adjacency matrix. AUC is defined as the area under the ROC curve, which is the indicator for a comprehensive evaluation of TPR (True Positive Rate) and FPR (False Positive Rate). The closer the AUC is to 1, the more clearly our model can tell whether there are edges or not.

\item[-] \textbf{ACC(net)}: We sample an estimated adjacency matrix $\hat{A}$ with a value of either 0 or 1 from the estimated edge probabilities. ACC(net) is the proportion of elements that correctly estimated of the adjacency matrix $\hat{A}$. And the ACC(net)-missing is the ACC(net) of the missing network structure. The values range from 0-1, and the closer the 1 value is, the better the result of the model is. 

\item[-] \textbf{ACC(states)}: For continuous time series data, ACC(states) refers to the MSE(Mean Square Error) between the predicted states $\hat{\bm{x}_t}$ and the ground truth states $\bm{x}_t$. While for discrete time series,  MAE(Mean Absolute Error) is obtained by predicting the state after sampling dispersed $\hat{x_t}$, ACC(states) of discrete time series equals $1-MAE$. In network completion tasks, the 
s are divided into two classes, one is observed nodes, the other is missing nodes. The states' accuracy of these two classes of nodes is called observed ACC(states), missing ACC(states) respectively. 

\item[-] \textbf{TPR}(True Positive Rate):
TPR measures the proportion of actual positives that are correctly identified in adjacent matrix $\hat{A}$. And the closer the TPR score is to 1, the lower the error rate is.

\item[-] \textbf{FPR}(False Positive Rate): FPR measures the proportion of actual positives that are wrongly identified in sampled adjacent matrix $\hat{A}$. The error rate decreases when the FPR score approaches 0. 

\end{itemize}

\subsection{Nodes alignment problem}
In the evaluation for network completion, because there are no nodes' labels, we need to find the matching between the nodes of missing and the ground truth to evaluate the effect of network completion. We used a greedy algorithm to find a nodes alignment. The main idea of the greedy algorithm is to compare the corresponding column or the row between $\hat{A}$ and $A$ and find the most similar vector correspondence. Here, a hamming distance is used to measure the similarity. By traversing the missing nodes, we can find the matching relationship between the estimate value and the ground truth of the missing nodes. The estimate of the rearranged adjacency matrix
is used for calculating all performance metrics.

\subsection{Experimental Setup}
\begin{table*}[h!]
\centering 
\resizebox{\textwidth}{!}{
\begin{tabular}{|c|c|c|c|c|c|c|c|c|c|c|}
  \hline
  \multirow{2}{*}{Data} & \multirow{2}{*}{Node Num} & \multicolumn{5}{c|}{GGN} & \multicolumn{4}{c|}{NRI}\\
  \cline{3-11}
  ~& &AUC&ACC(net)&TPR&FPR&ACC(state)&ACC(net)&TPR&FPR&ACC(state)\\
  \hline
  \multirow{3}{*}{Voter} &10 & 0.987 & 0.952 & 0.930 & 0.033 & 0.928 & 0.666 & 0.530 &0.241&0.920\\
  \cline{2-11}
  ~&20 & 0.997 & 0.989 & 0.982 & 0.009 & 0.867 & 0.750 & 0.559&0.198&0.863\\
  \cline{2-11}
  ~&30 & 0.999 & 0.996 & 0.976 & 0.001 & 0.870 & - & -& -&-\\
  \hline
  \multirow{3}*{CML}&10 & 1 & 1 & 1 & 0 & 5.63E-06 & 0.531 & 0.446 & 0.588&1.69E-04\\
  \cline{2-11}
  ~&20 & 1 & 1 & 1 & 0 & 3.00E-06 & - & - & - &- \\
  \cline{2-11}
  ~&30 & 1 & 1 & 1 & 0 & 3.29E-06 & - & - & - &- \\
  \hline
  
\end{tabular}}
\caption{Network Reconstruction performance on Voter and CML data }
\label{nrresult}
\end{table*}
The parameter settings on our three modules are as follows: (1) In the dynamics learner module, we use 4-layered MLP as a function of information aggregation between nodes. The activation function of each layer is ReLU. The model parameter $\alpha $ is a randomly initialized set of parameters. When the number of neurons in the hidden layers is set to 64, 32, 16, 8, respectively, the network reconstruction task can achieved good results. As to the network completion task, on the Voter dataset, the settings are similar. On the CML dataset, the embedding dimension of hidden layer is set to 32 dimensions, and that of the 2-4 layers is 16, 8, and 4 dimensions, respectively. Within each epoch, we trained the dynamics learner module 30 times. (2) In the initial state learning module of the network completion task, we use different methods to deal with it. In the case of discrete state, the nodes' states are usually coded by one-hot vectors to indicate which kind of state the node belongs to. We use sigmoid as the function $\rho$ to classify the parameters into the interval $[0,1]$, indicating the probability that the node belongs to a certain state. In the case of continuous states, the state has no interval limit, and the function $\rho$ is the identity mapping of the parameter $\gamma$. On the CML dataset, we find that we can achieve similar completion accuracy without updating the initial states of unknown nodes when updating the missing initial states, so we set the initial states as random values in the training process. (3) In the network generator module, we randomly generate an initial set of learnable parameters from the normal distribution $N(0,1)$ for training.The above three modules are optimized using the Adam algorithm, and the learning rate is $0.001,0.1,0.1$ respectively.

Since the initial states learning module in the network completion task only learned the initial states of the unknown nodes of the training set, but not the states of the unknown nodes of the test set. Therefore, during the test phase, we fix the learned dynamics learner and adjacency matrix to generate a set of trainable unknown nodes' states for the test set. Similar to the initial states learning module of the training set, we optimize the initial state of the unknown nodes in the test set and calculate the accuracy of the dynamic predictor.

\subsection{Inference Accuracy}

\subsubsection{Accuracy of inference with different network sizes}
Our approach is compared with the neural relational inference model on the task of network reconstruction.
\begin{itemize}
\item[-] \textbf{NRI}(Neural Relational Inference Model) applies a variational auto-encoder method to learn the underlying interaction graph and the complex system dynamics from the observational dynamical data. We ran the NRI Model on the Voter dataset and the CML dataset using settings consistent with the original paper of Kipf etc \cite{kipf2018neural}.  
\end{itemize}

As to the comparative model for network completion task, according to our investigation, there are almost no models can be applied directly to solve the network completion based on time series data. At the same time, modifying other types of existing models for network completion will be a very complicated problem. Therefore, in the network completion tasks, we do not conduct comparative experiments.We demonstrate the accuracy of the network inference model across all data sets on different network sizes.

In table \ref{nrresult}, we show the performances of the GGN and NRI model on the network reconstruction task.
\begin{table*}[h!]
\centering
\resizebox{\textwidth}{!}{
\begin{tabular}{|c|c|c|c|c|c|c|c|}
  \hline
   \multirow{2}*{Data} & \multirow{2}{*}{Num-Missing Num}& \multicolumn{6}{c|}{NC-GNN}\\
  \cline{3-8}
  ~& &Missing AUC & Missing ACC(net) & TPR & FPR & Missing ACC(states) & Observed ACC(states) \\
  \hline
  \multirow{2}*{Voter} &10-1 &0.849 & 0.724 & 0.719 & 0.276 & 0.895 & 0.942\\
  \cline{2-8} 
  ~&20-2& 0.732 & 0.670 & 0.615 & 0.347 & 0.791 & 0.914 \\
  \hline
  \multirow{2}*{CML} &10-1 & 0.993&0.867 & 0.809 & 0.089 & - & 0.009 \\
  \cline{2-8} 
  ~&20-2 & 0.803& 0.786 & 0.827 & 0.242 & - & 0.066\\
  
  \hline
\end{tabular}}
\caption{Network Completion performance comparison on Voter and CML data}
\label{ncresult}
\end{table*} We empirically show that the average metrics scores in the experiments on three scales. In terms of the accuracy of structure inference, AUCs can reach above 98\% and ACC above 95\% in networks of different sizes. The values of TPR and FPR of the reconstructed adjacency matrix are close to the optimal values. The accuracy of ACC(states) is over 86\% in the Voter data set. The error rate (MAE) is close to 0 in CML data sets, indicating that GGN can better fit the dynamics of the network. In addition, it can be also observed that GGN performs better than the NRI model on network reconstruction tasks. 

We carry out the experiment of the NC-GGN model on network completion on the same data set, where the percent of missing nodes $M$ is set to 10\%. In table \ref{ncresult}, the first number in the first column represents the total network size, and the second number is the number of nodes being removed. For example, 20-2 indicates that there are 20 nodes in a complete dynamical system, and only 18 nodes are observed. The information on the two nodes was completely missing.

\begin{figure*}[ht!]
\centering
\includegraphics[scale=0.7]{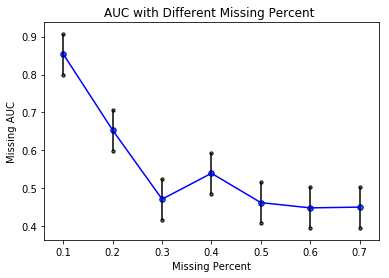}
\caption{AUC decreases with the proportions of missing nodes.}
\label{ncmissingresult}
\end{figure*}
Both in the inference of missing structure and nodes' states, our model NC-GGN has achieved high inference accuracy on a 10-node scale network. As the scale of the network increases, the number of possible connections increases greatly, and the difficulty of completion also increases, as a result, the accuracy decreases. The accuracy of Voter data sets is lower than that of the CML data sets, which is due to the fact that the number of CML samples is more than that of the Voter data sets. We can also see that the accuracy of the network completion task is lower than that of the network reconstruction task. 

\subsubsection{Accuracy with different missing proportions}

In this part, we investigate the effect of the observed network completeness on network completion problem. We adjust the completeness of the network by tuning the proportion of missing nodes. The higher the proportion of missing nodes is, the less complete the system is. We create a number of partially observable network $G_o$ with 10\% to 70\% missing Nodes on CML data sets based on 20-node WS networks. We plot the AUCs with the missing proportions in figure \ref{ncmissingresult}.

We mentioned that before the proportion of missing nodes is increasing, and the difficulty of network completion increases gradually. When the proportion of missing nodes changes from 0.1 to 0.3, the performance of completion decreases significantly. GGN fails to complete the network if the nodes' missing proportion exceeds 0.3 because the AUC values are indistinguishable with the random guess.

\section{Discussion}
In this paper, we take the form of Graph Neural Network to solve two types of Network Inference problems: Network Reconstruction and Network Completion. First, we formulate the network inference problems based on time series data as the optimization problems. Second, in the model part, we extended our previous framework GGN to NC-GGN framework for network completion, which can be applied to a variety of dynamical time series data. Third, in the experimental part, we demonstrated GGN model can reconstruct network structure accurately based on time series data without any prior knowledge of network structure. Meanwhile, NC-GGN can infer the hidden node's states and structural information from 90\% of known network structure and dynamical data. Besides, the performance of network completion task is influenced by the proportion of missing nodes seriously.

There are still many aspects that can be improved in our current work. For example, all of the experiments are carried out on small-sized networks with less than 30 nodes. In the future, the network scale can be increased from two aspects by increasing the computational power or by simplifying the model. Besides, the amount of data needed for network inference is relatively large, the reduction for data requirement will affect the inference performance of the model. Furthermore, the result of network completion can be further improved by increasing the accuracy of the missing initial states.
\section*{ACKNOWLEDGMENTS}
The research is supported by the National Natural Science Foundation of China(NSFC) under the grant numbers 61673070.

\bibliographystyle{IEEEtran}
\bibliography{main}

\end{document}